\documentclass[conference]{IEEEtran}
\IEEEoverridecommandlockouts

\usepackage{cite}
\usepackage{amsmath,amssymb,amsfonts}
\usepackage{algorithm}
\usepackage{algpseudocode}
\usepackage{graphicx}
\usepackage{textcomp}
\usepackage{xcolor}

\ifCLASSOPTIONcompsoc
\usepackage[caption=false,font=normalsize,labelfon
t=sf,textfont=sf]{subfig}
\else
\usepackage[caption=false,font=footnotesize]{subfi
g}
\fi

\def\BibTeX{{\rm B\kern-.05em{\sc i\kern-.025em b}\kern-.08em
    T\kern-.1667em\lower.7ex\hbox{E}\kern-.125emX}}
\begin{document}

\bstctlcite{IEEEexample:BSTcontrol}

\title{Variable-Frequency Imitation Learning for Variable-Speed Motion
\thanks{This work was supported by JSPS KAKENHI Grant Number 24K00905, JST, PRESTO Grant Number JPMJPR24T3 Japan, and JST ALCA-Next Japan, Grant Number JPMJAN24F1. This study was based on the results obtained from the JPNP20004 project subsidized by the New Energy and Industrial Technology Development Organization (NEDO).}
}

\author{\IEEEauthorblockN{Nozomu Masuya}
\IEEEauthorblockA{\textit{Grad. School of Science and Technology} \\
\textit{University of Tsukuba}\\
Tsukuba, Japan \\
masuya.nozomu.sm@alumni.tsukuba.ac.jp}
\and
\IEEEauthorblockN{Sho Sakaino}
\IEEEauthorblockA{\textit{Systems and Information Engineering} \\
\textit{University of Tsukuba}\\
Tsukuba, Japan \\
sakaino@iit.tsukuba.ac.jp}
\and
\IEEEauthorblockN{Toshiaki Tsuji}
\IEEEauthorblockA{\textit{Grad. School of Science and Engineering} \\
\textit{Saitama University}\\
Saitama, Japan \\
tsuji@ees.saitama-u.ac.jp}
}

\maketitle

\begin{abstract}
    Conventional methods of imitation learning for variable-speed motion have difficulty extrapolating speeds because they rely on learning models running at a constant sampling frequency. 
    This study proposes variable-frequency imitation learning~(VFIL), a novel method for imitation learning with learning models trained to run at variable sampling frequencies along with the desired speeds of motion. 
    The experimental results showed that the proposed method improved the velocity-wise accuracy along both the interpolated and extrapolated frequency labels, in addition to a 12.5~\% increase in the overall success rate.
\end{abstract}


\section{Introduction}
    The demand for human-like, cooperative, adaptive, and quick robot manipulation has increased, especially in fields suffering from a shortage of workers. 
    To achieve such manipulation of robots without hard coding, machine learning-based methods are gaining attention. 
    Reinforcement learning-based methods~\cite{levine2018, daydreamer} are completely free of prior modeling and teaching. These methods have proven to be effective in real-world manipulation tasks, although at the cost of enormous trial and error, which causes wear in robots.
    By contrast, imitation learning~\cite{yang2017, aloha} has enabled robots to mimic human motion from human demonstrations, with machine learning models generating command values in accordance with sensory data. 

    Learning motions at variable-speed settings contributes to a more precise understanding of the dynamics of robots and the surrounding environment. Despite the recent shift toward world models~\cite{guan2024, mitrokhov2024, daydreamer} for artificial general intelligence, including general robot manipulation, model-free imitation learning methods for generating variable-speed motions suffer from difficulties in extrapolating the speeds because the variable-speed task is perceived as a multi-task setting.

    This study proposes variable-frequency imitation learning~(VFIL), a model-free imitation learning method that considers variable-speed tasks as single-task settings by changing the sampling frequency of the neural network~(NN) models according to the frequency command of the ongoing motion.
    The proposed method was experimentally verified on a contact-rich variable-speed wiping task.
    The results showed that the proposed method has a high accuracy in terms of speed and a high success rate, especially in high-frequency settings.

\section{Related Works}
    \subsection{Extraction of Skills from Variable-Speed Demonstrations}
        Time-series alignment has long been a field of research to analyze time-series data more concisely. In robotics, the extraction of skills from demonstrations is one of its use cases.

        Cohen~\textit{et al.} proposed Gromov’s dynamic time warping~\cite{cohen2021}, which is an extension of dynamic time warping~\cite{sakoe1978}, and verified it in imitation learning settings using time-series generative adversarial networks.
        In addition, Rasines~\textit{et al.} extended dynamic time warping and generated smooth trajectories from datasets obtained from real-world robots~\cite{rasines2023}, whereas the generated trajectories were not verified in an actual environment. 

        Methods based on dynamic movement primitives~\cite{saveriano2023} are also used for learning from demonstrations using a variety of time series. 
        Xu~\textit{et al.} achieved a shorter execution time and lower acceleration in a water-carrying task through nonlinear speed scaling~\cite{XU2024}.

        Although these approaches aim to extract skills from demonstrations with different timings, they do not focus on generating motion at variable speeds or adaptation on a real-time basis with a feedback loop. 
        This can lead to limited adaptability to variable-speed dynamic motions whose commands can vary according to their frequencies. 

    \subsection{Generation of Variable-Speed Motion from Demonstrations}
        Although humans can easily adjust the speed of motion of everyday activities, including handling unknown objects or conducting contact-rich tasks, a method to achieve variable-speed motion in hard-to-simulate robot manipulation has not been developed.

        Under assumptions including negligible inertial forces, Perico~\textit{et al.} achieved variable-speed motion through probabilistic principal component analysis~\cite{perico2020} in the settings of learning from demonstrations~\cite{perico2019}. 
        Although this method can generate motion at variable settings using relatively digestible algorithms, the assumptions behind the method render it inapplicable to high-speed motion, including that at high accelerations.

        In addition, under the assumption that environmental changes are negligible, variable-speed teaching playback with position-force control developed by Yokokura~\textit{et al.}~\cite{yokokura2009, fujisaki2023} can also be regarded as the achievement of variable-speed motion from demonstrations.
        
        Without these assumptions, Sakaino~\textit{et al.} verified the feasibility of variable-speed motion in imitation learning by using neural network~(NN) models and force control~\cite{sakaino2022}. 
        Saigusa~\textit{et al.} implemented self-supervised learning using this method~\cite{CRANEX7params} to achieve time-wise accuracy. 
        Although the fine-tuning method contributed to improved accuracy for interpolations, extrapolations to unseen speeds remain to be achieved, with their NN models trained at a constant sampling frequency, regardless of the frequency of the robots' actions. 
        This study distinguishes itself by changing the sampling frequency of the NN model according to the command value of the frequency of motion as a velocity-wise inductive bias.

        In addition, in the context of data augmentation, velocity-wise transformations are considered to increase the diversity of datasets.
        The model proposed by Yamamoto~\textit{et al.} could select various objects from a conveyor belt moving at variable speeds by diversifying the speed and phase of the motion data obtained from three human demonstrations via downsampling~\cite{yamamoto2023}. 
        Although this method achieved end-to-end motion generation by adapting to a variable-speed environment, it did not focus on changing the speed in accordance with external commands.

\section{Method: Variable-Frequency Imitation Learning}
\begin{figure}[tb]
        \centering
        \subfloat[Conventional method]{%
            \resizebox*{8cm}{!}{\includegraphics[scale=0.6, bb = 0 0 788 407]{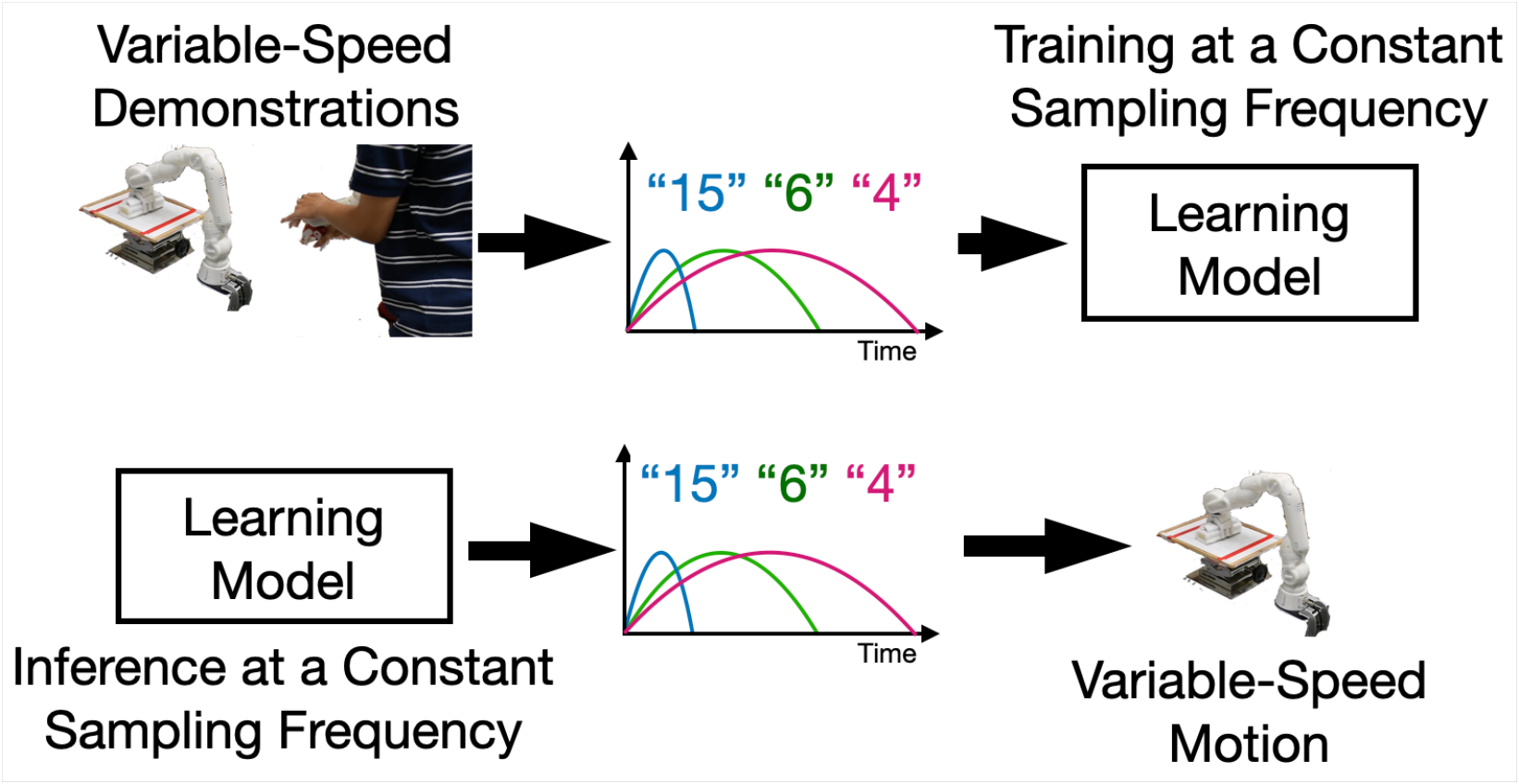}}
            \label{conventional}
        }
        \vspace{1pt}
        \centering
        \subfloat[Variable-frequency imitation learning]{%
            \resizebox*{8cm}{!}{\includegraphics[scale=0.6, bb = 0 0 775 779]{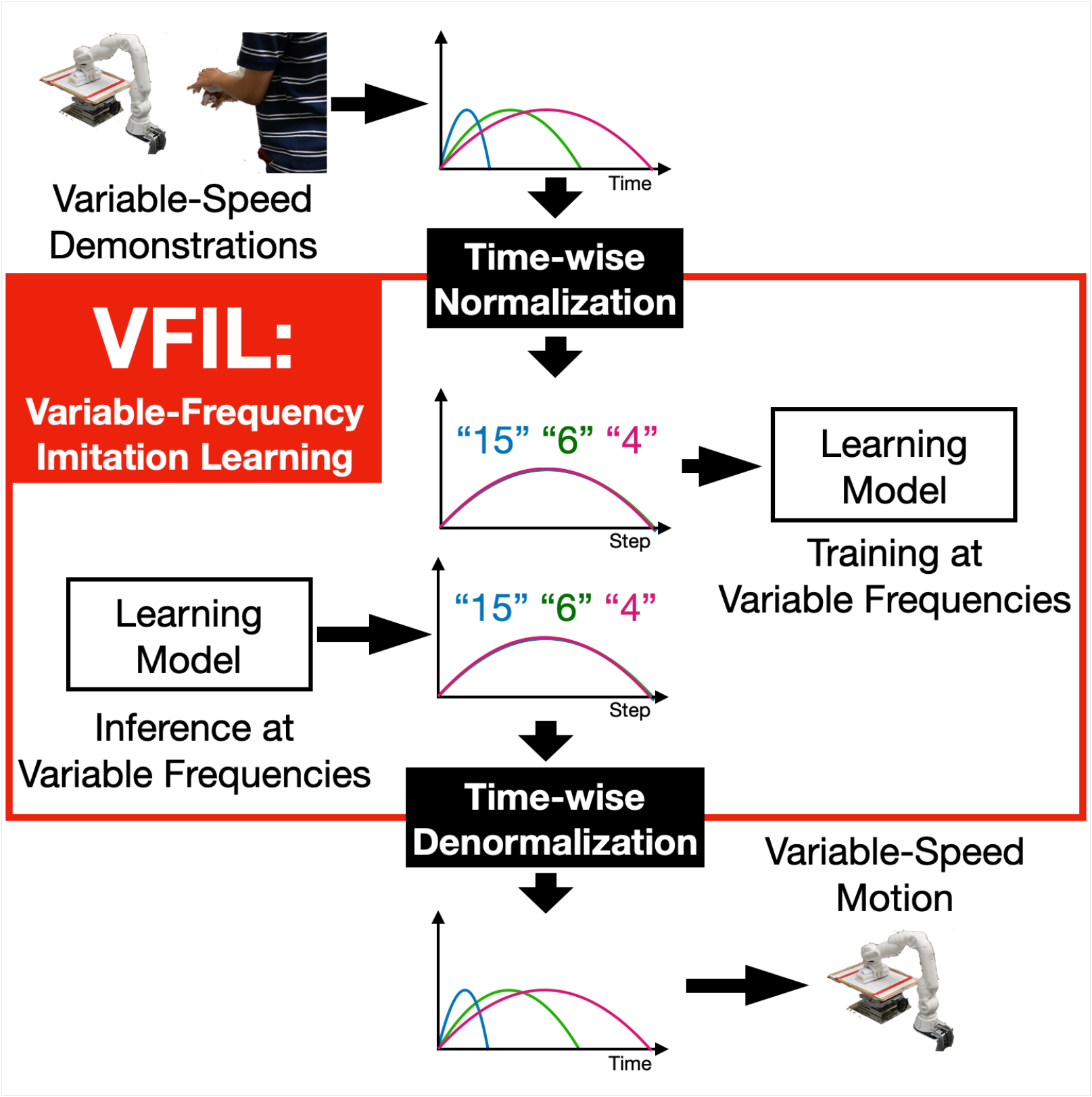}}
            \label{proposal}
        }
        \caption{Flow of the conventional and the proposed methods.}
        \label{fig:proposal}
    \end{figure}

    This study proposes variable-frequency imitation learning~(VFIL), a novel method of imitation learning with time-wise normalization and learning models running at variable frequencies.
    The flow of the VFIL is shown in Fig.~\ref{fig:proposal}.

    In conventional imitation learning methods, the sampling frequencies of the learning models remain unchanged between the training data. 
    When learning variable-speed motions, trajectories are acquired at various frequencies, and motion is learned at a constant sampling frequency, forcing the learning model to adapt to environmental changes and generate different motions at variable speeds.
    This complicates the learning process as simple changes in speed convert the problem of learning variable-speed motion into multi-task learning problems, leading to poor performance, especially in extrapolations.
   
    In contrast, through time-wise normalization, the proposed method treats the changes in the time domain and the related nonlinearity, equally as changes in environments and robot systems, such as rough surfaces, heavy frames, or worn gears.
    Under this inductive bias, the model simply learns to be robust against nonlinear changes, being free from following multiple trajectories in time series.

    The pseudo-codes for time-wise normalization and denormalization are shown in Algorithms~\ref{alg1} and \ref{alg2}.
    In this method, $N$ demonstrations of training data with a sampling frequency $F$ and a variety of motion frequencies $f_1, f_2, \cdots, f_N$ were resampled to a constant motion frequency $f_0$ with labels indicating the original speeds.
    To minimize the obvious differences between the training data, the velocity data were scaled by factors of $f_0 / f_1, f_0 / f_2, \cdots f_0 / f_N$.
    This time-wise normalization enables the training of the learning model at different sampling frequencies, $f_1 F / f_0, f_2 F / f_0, \cdots f_N F/ f_0$, for each demonstration with different environmental reactions according to the frequency labels $f_1, f_2, \cdots, f_N$.
    The inference was conducted in real time according to the desired motion frequency $f$.
    The learning model was given the frequency label $f$ and was run at the average sampling frequency of $fF/f_0$ in addition to the velocity data scaled at $f/f_0$ as time-wise denormalization.

\begin{figure}[tb]
        \centerline{\includegraphics[scale=0.25, bb=0 0 904 441]{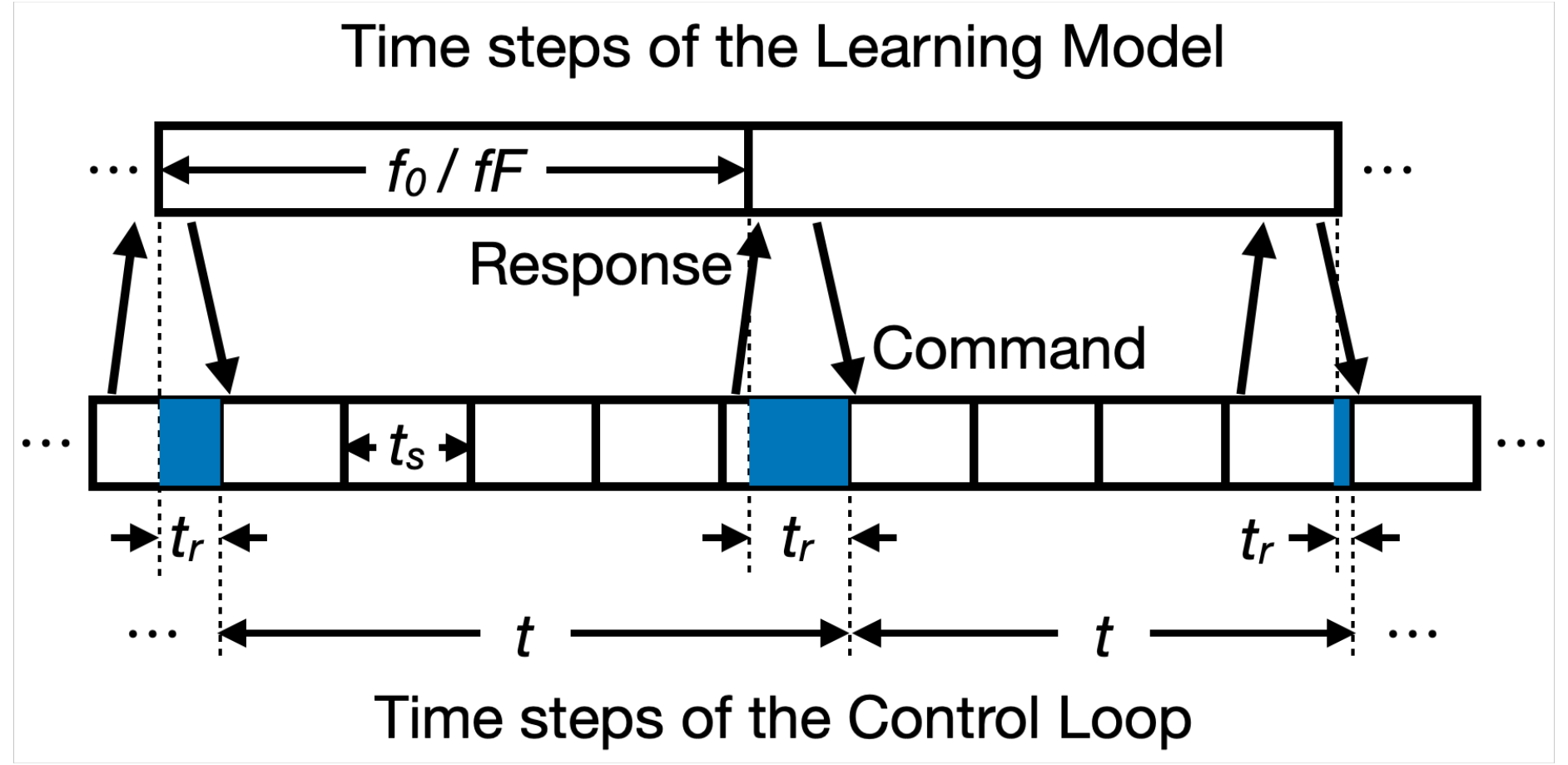}}
        \caption{Adjustment of time steps between the learning model and control loop.}
        \label{fig:contloop}
    \end{figure}

    To preserve the control system across all variable-frequency motions, the sampling rate of the control system was kept constant and higher than that of the learning model. 
    Fig.~\ref{fig:contloop} depicts the adjustment of the sampling times between the learning model and the control loop.
    In the inference phase, the progression of the learning model to the next step is determined at each step of the control system. 
    The learning model moves to the next step when the sum of the elapsed time from the model's last step $t$ and the remainder from the second last step $t_r$ exceeds $f_0/fF$.
    The remaining time $t - f_0/fF$ is passed on to the nest step as $t_r$.
    To calculate the time required for the learning model, the inputs and outputs were obtained at different time steps.

    \begin{figure}[tb]
        \begin{algorithm}[H]
            \caption{Time-series normalization}
            \label{alg1}
            \begin{algorithmic}[1]
            \State $f_0 \gets base\ motion\ frequency$
            \ForAll{demonstration data with motion frequency $f_i$}
                \State \textbf{Resample} demonstration at $(f_0 / f_i)$ times the speed
                \State $ velocity \gets velocity \times (f_0 / f_i)$
            \EndFor\\
            \Return{training data}
            \end{algorithmic}
        \end{algorithm}
    \end{figure}

    \begin{figure}[tb]
        \begin{algorithm}[H]
            \caption{Time-series denormalization along with the control loop}
            \label{alg2}
            \begin{algorithmic}[1]
            \State \textbf{Initialize} $t, t_r$
            \State $t_s \gets control\ loop's\ sampling\ time$
            \State $F \gets learning\ model's\ frequency$
            \State $f_0 \gets base\ motion\ frequency$
            \State $f \gets task\ motion\ frequency$
            \While{the task is ongoing}
                \State $(\bold{\theta}^{res}, \bold{\omega}^{res}, \bold{\tau}^{res}) \gets robot's\ response$
                \Comment (angle,~angular~velocity,~torque)
                \If{$t + t_r \geq f_0/fF$}
                    \State $ \bold{\omega}^{res} \gets \bold{\omega}^{res} \times (f_0 / f)$
                    \State \textbf{send} $\bold{\theta}^{res}, \bold{\omega}^{res}, \bold{\tau}^{res}$ to the learning model
                    \State $t_r \gets t - f_0/fF$
                    \State $t \gets 0$
                \EndIf
                \If{the next step of sending}
                    \State $(\bold{\theta}^{cmd}, \bold{\omega}^{cmd}, \bold{\tau}^{cmd}) \gets robot's\ command$
                    \Comment (angle,~angular~velocity,~torque)
                    \State $\bold{\omega}^{cmd} \gets \bold{\omega}^{cmd} \times (f / f_0)$
                    \State \textbf{update} the command value with $\bold{\theta}^{cmd}, \bold{\omega}^{cmd}, \bold{\tau}^{cmd}$
                \EndIf
                \State \textbf{Calculate} the control input
                \State \textbf{Input} the control input to the robot
                \State $t \gets t + t_s$
                \State \textbf{wait} for the remainder of the control cycle
            \EndWhile
            \end{algorithmic}
        \end{algorithm}
    \end{figure}

\section{Experiments and Evaluations}
    We evaluated the proposed method using bilateral control-based imitation learning~\cite{adachi2018, CRANEX7params} along with the conventional method with a constant sampling frequency.

\begin{figure}[tb]
        \centering
        \subfloat[Four-channel bilateral control]{%
            \resizebox*{6cm}{!}{\includegraphics[scale=0.5, bb = 0 0 718 324]{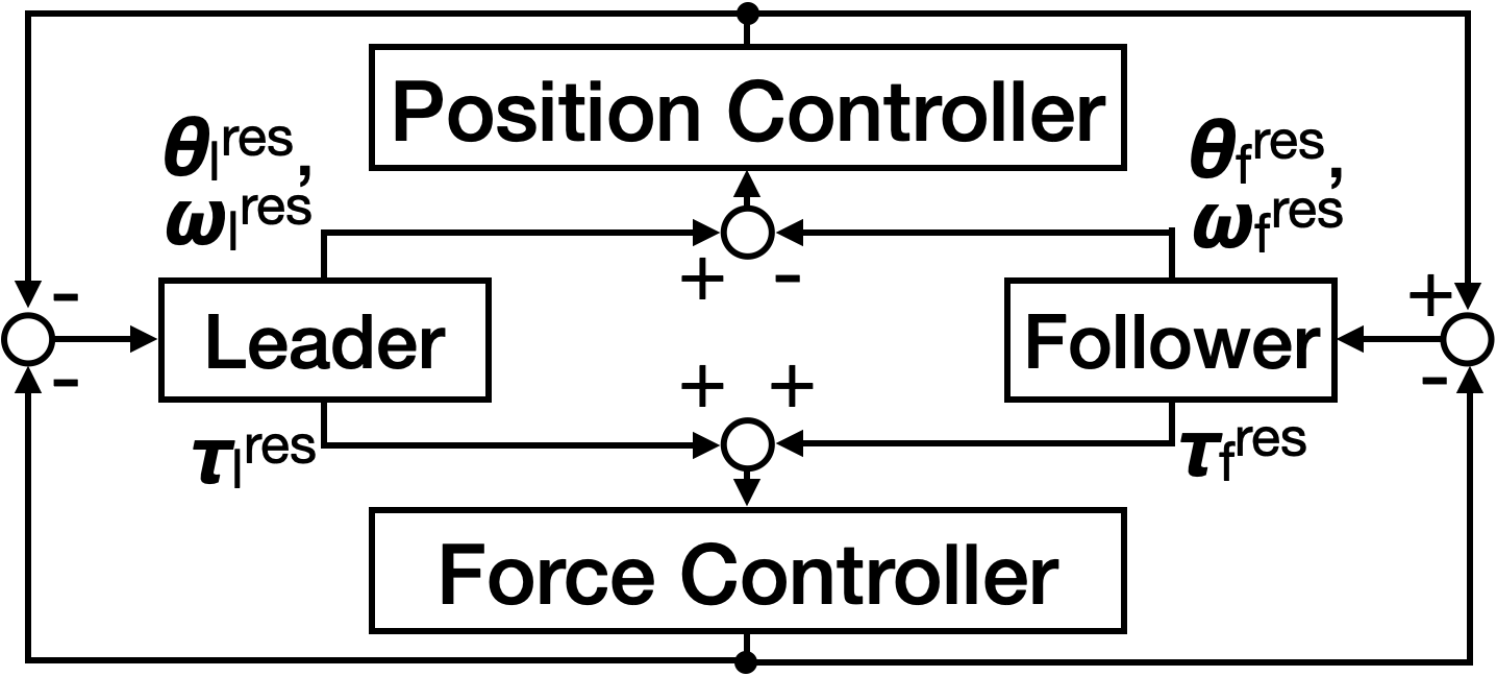}}
            \label{4ch}
        }
        \vspace{1pt}
        \subfloat[Bilateral control-based imitation learning]{%
            \resizebox*{6cm}{!}{\includegraphics[scale=0.5, bb = 0 0 704 315]{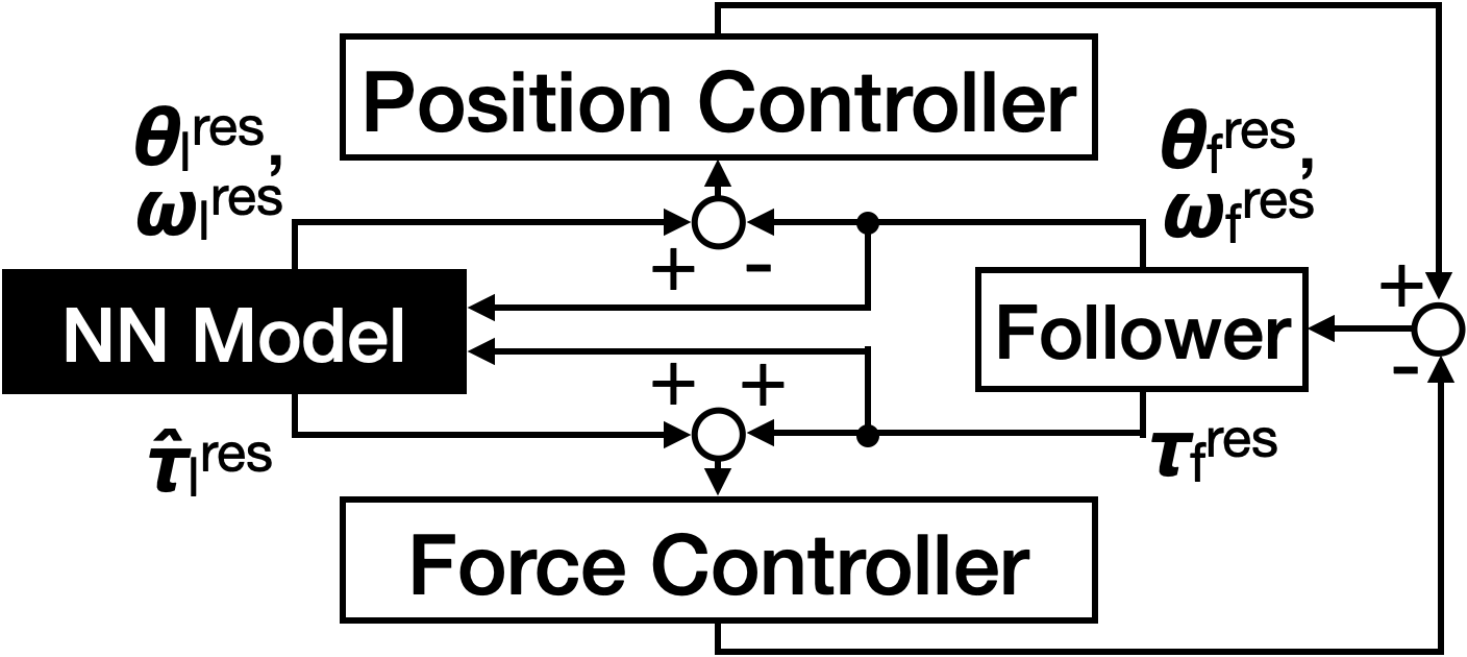}}
            \label{moho}
        }
        \caption{Block diagrams of four-channel bilateral control and bilateral control-based imitation learning.}
        \label{fig:4chmoho}
    \end{figure}

    In bilateral-control-based imitation learning, human demonstrations are collected through four-channel bilateral control~\cite{sakaino2011}, which is a teleoperation method with symmetrical position-force feedback.
    The objective of the four-channel bilateral control is to satisfy the following equations:
    \begin{equation}
        \bold{\theta}^\mathrm{res}_f - \bold{\theta}^\mathrm{res}_l = \bold {0} 
    \end{equation}
    and
    \begin{equation}
        \bold{\tau}^\mathrm{res}_f + \bold{\tau}^\mathrm{res}_l = \bold {0}, 
    \end{equation}
    between the leader robot on the operator’s side and the follower robot physically conducting the task.
    A block diagram of the four-channel bilateral control system is shown in Fig.~\ref{4ch}.
    In both the equations and block diagrams, the variables $\bold{\theta}$, $\bold{\omega}$ and $\bold{\tau}$ denote the vectors of the joint angle, angular velocity, and torque, respectively.
    The superscript $ref$ indicates the reference value, and $res$ indicates the response value.
    The subscripts $l$ and $f$ refer to the leader and follower, respectively.

    After training the NN model, the motion was reproduced using the NN model instead of the leader robot, as shown in Fig.~\ref{moho}.
    The circumflex represents the value estimated using the NN model.

    \subsection{Experimental Setup}
        \subsubsection{Setup of the Task}
\begin{figure}[tb]
                \centering
                \subfloat[Initial state]{%
                    \resizebox*{8cm}{!}{\includegraphics[scale=0.15, bb = 0 0 1920 1080]{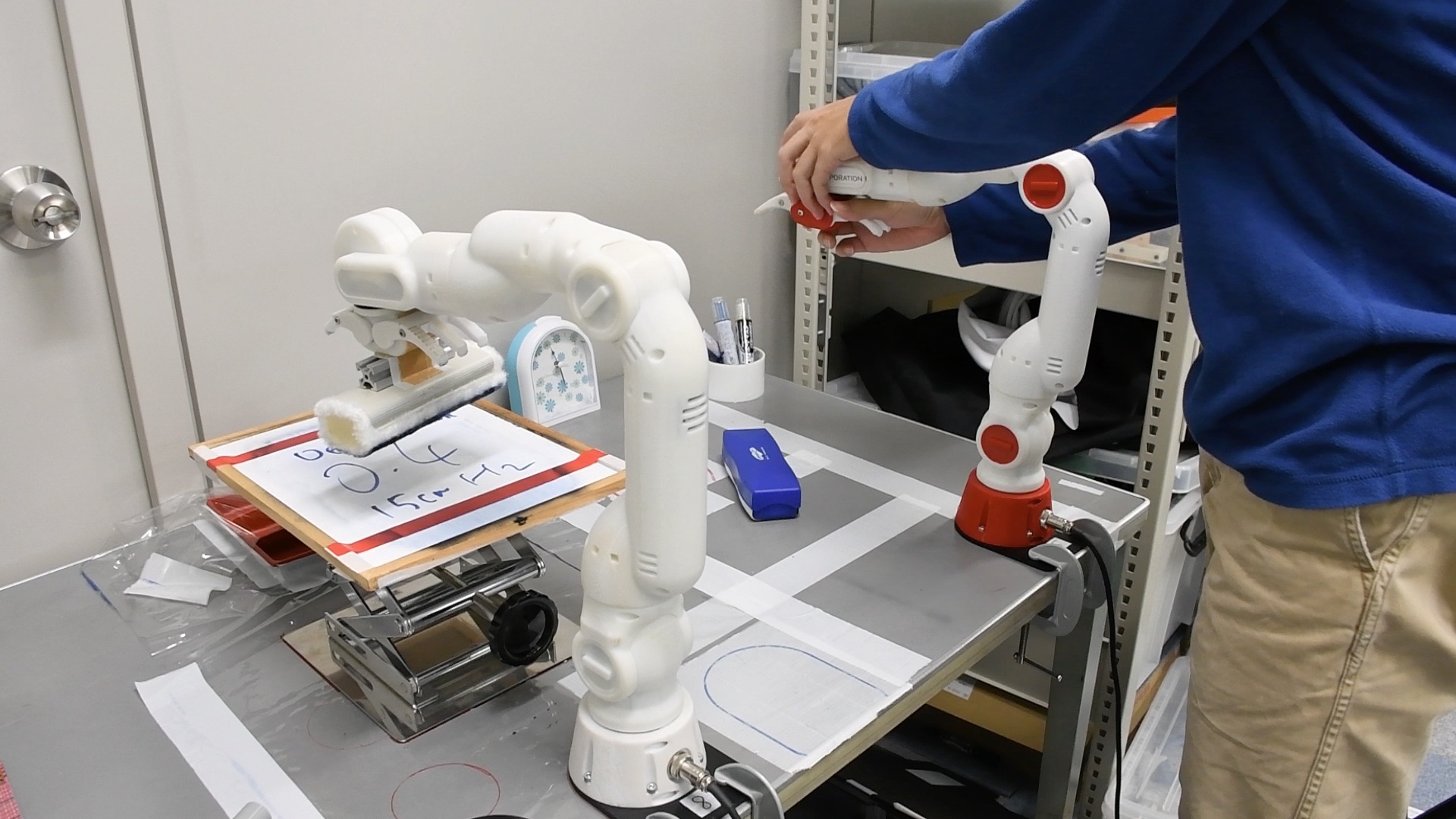}}
                }
                \hspace{5pt}
                \subfloat[Pressing]{%
                    \resizebox*{8cm}{!}{\includegraphics[scale=0.15, bb = 0 0 1920 1080]{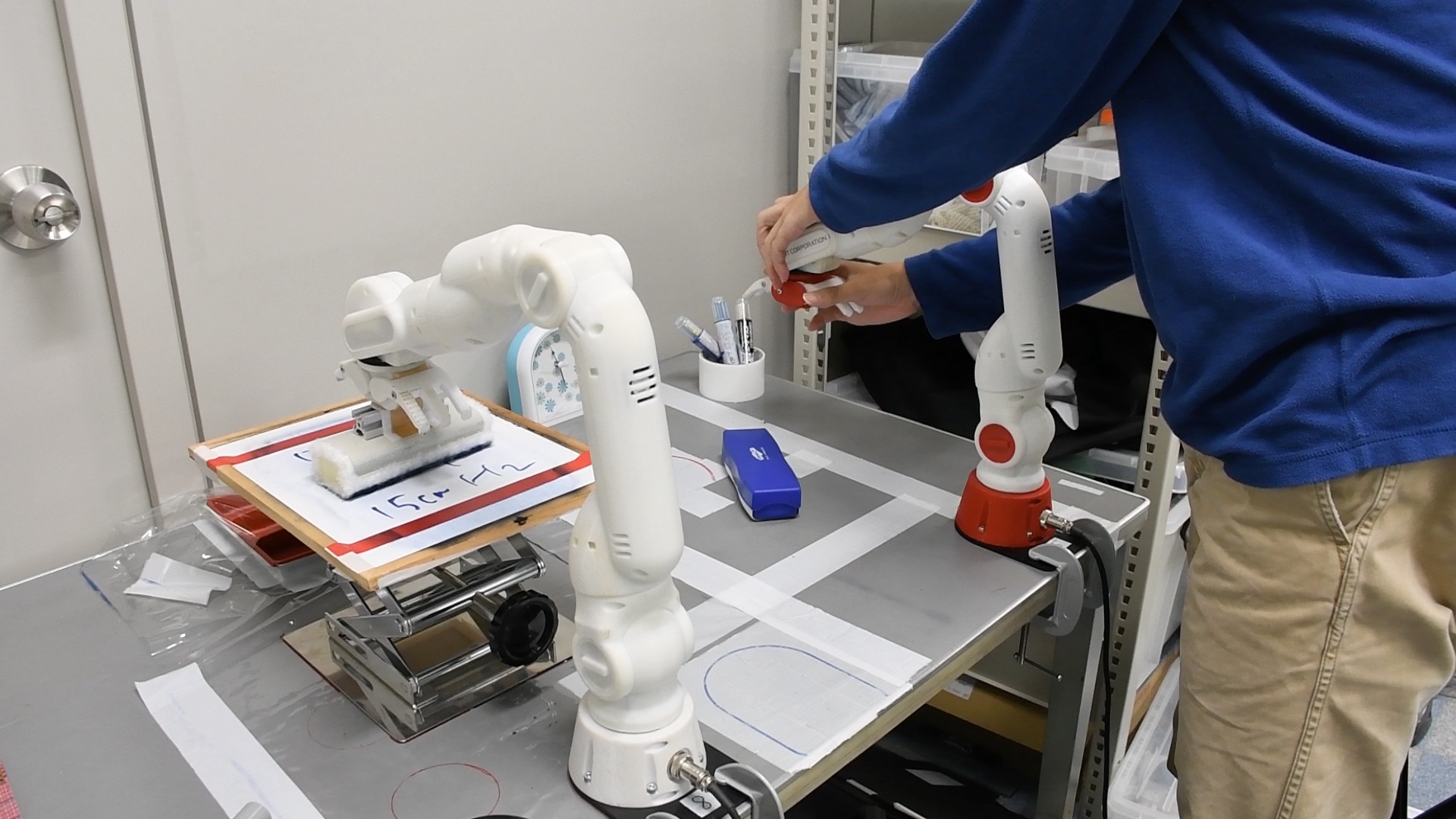}}
                }
                \hspace{5pt}
                \subfloat[Grasping]{%
                    \resizebox*{8cm}{!}{\includegraphics[scale=0.15, bb = 0 0 1920 1080]{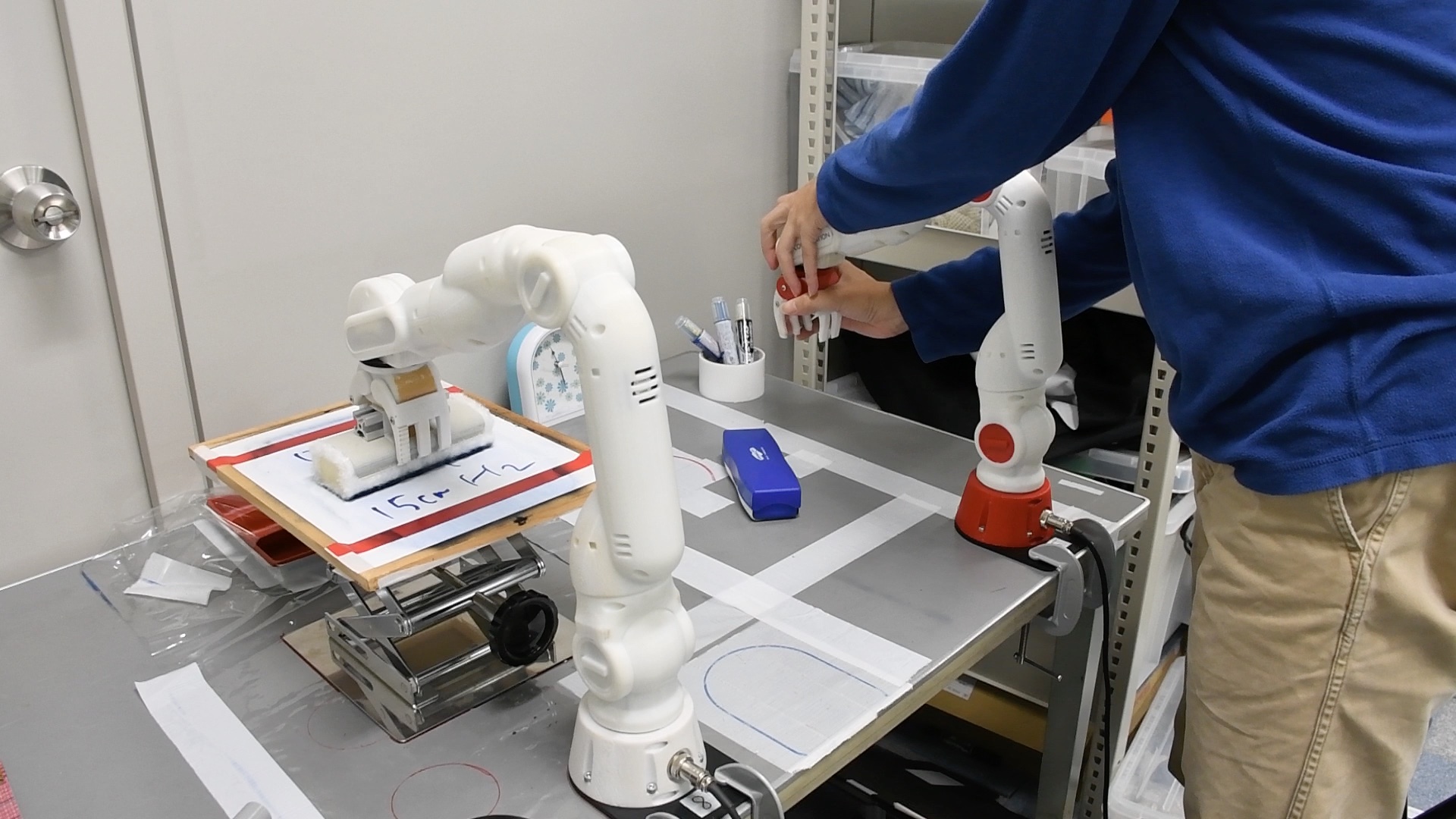}}
                }
                \hspace{5pt}
                \subfloat[Wiping]{%
                    \resizebox*{8cm}{!}{\includegraphics[scale=0.15, bb = 0 0 1920 1080]{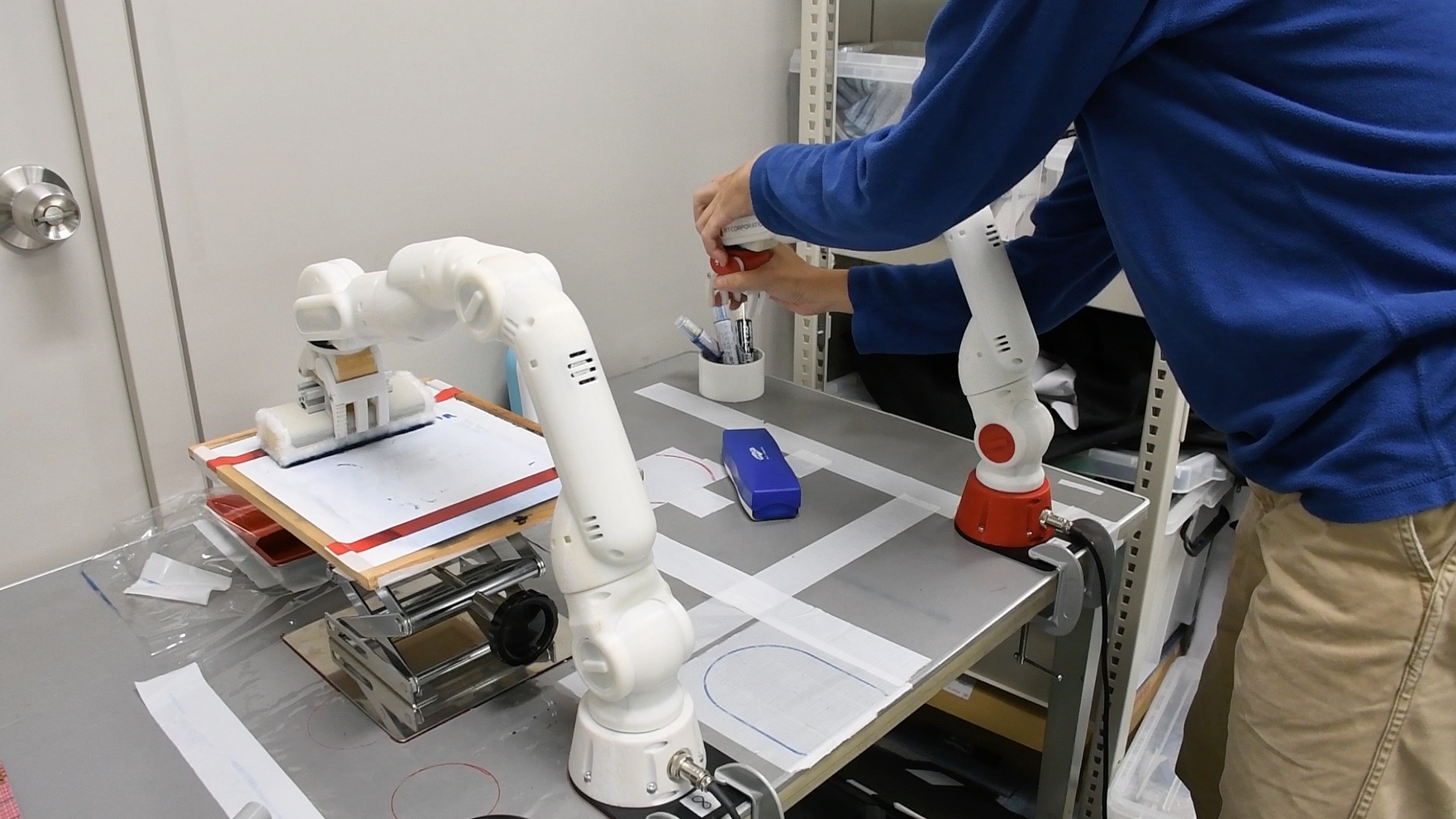}}
                }
                \caption{Procedure of the wiping task taught through bilateral control.}
                \label{fig:task}
            \end{figure}

            To evaluate the stability and adaptability of the proposed method, we prepared the setup for a wiping task, as shown in Fig.~\ref{fig:task}.
            In this task, the robot must first detect the height of the wiping surface by pressing the eraser onto the whiteboard. It should then grasp the eraser and wipe the board continuously at a prescribed frequency.
            Each trial lasted 40 s, including the pressing, grasping, and wiping tasks.
            A trial was considered to have failed if the eraser lost contact with the whiteboard at any time after the pressing phase.

            In the teaching process, 18 demonstrations were collected through teleoperation, three times each at 0.4, 0.6, and 0.8~Hz and at heights of 10 and 15~cm.
            To collect data at constant frequencies, demonstrations were conducted with the metronome set at 48, 72, and 96 beats per minute, moving either forward or backward at each beat.

            In the trials, the commands for the wiping frequencies were set to the unlearned 0.2, 1.0, and 1.4~Hz, in addition to the learned 0.6~Hz.

        \subsubsection{Setup of Robots}
\begin{figure}[tb]
                \centerline{\includegraphics[scale=0.064, bb=0 0 2647 2118]{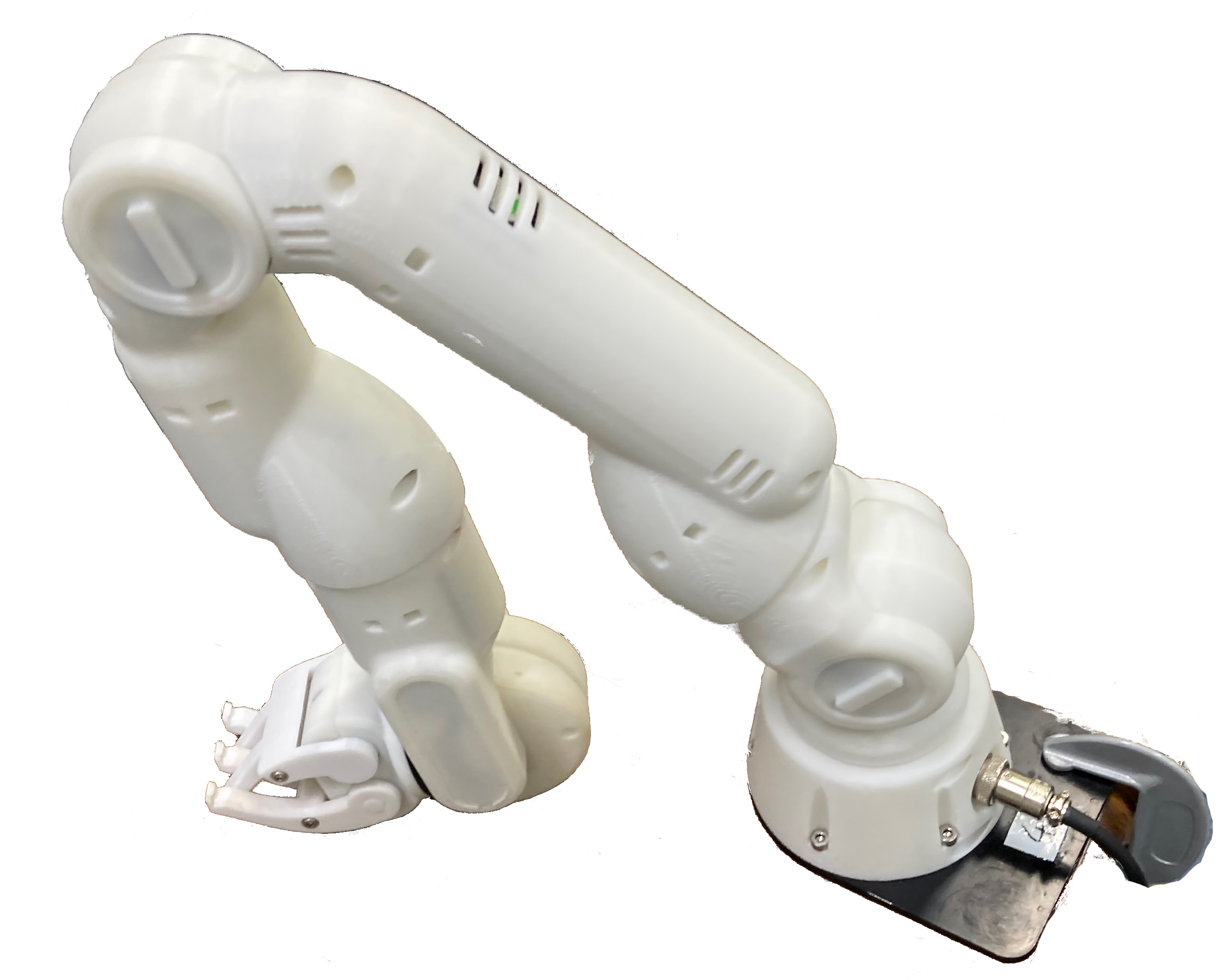}}
                \caption{CRANE-X7 manipulator with a cross-structured hand.}
                \label{fig:crane}
            \end{figure}

            We used two CRANE-X7 manipulators~(RT Corp., Tokyo, Japan), as shown in Fig.~\ref{fig:crane}.
            Each manipulator had an arm with seven degrees of freedom (DoF) and a rigid end effector with one DoF.
            The third joint of the arms was constrained to eliminate the redundant DoFs and the end effectors were replaced with the cross-structured hand developed by Yamane \textit{et al.}~\cite{yamaneral}.

\begin{figure}[tb]
                \centerline{\includegraphics[scale=0.26, bb=0 0 966 319]{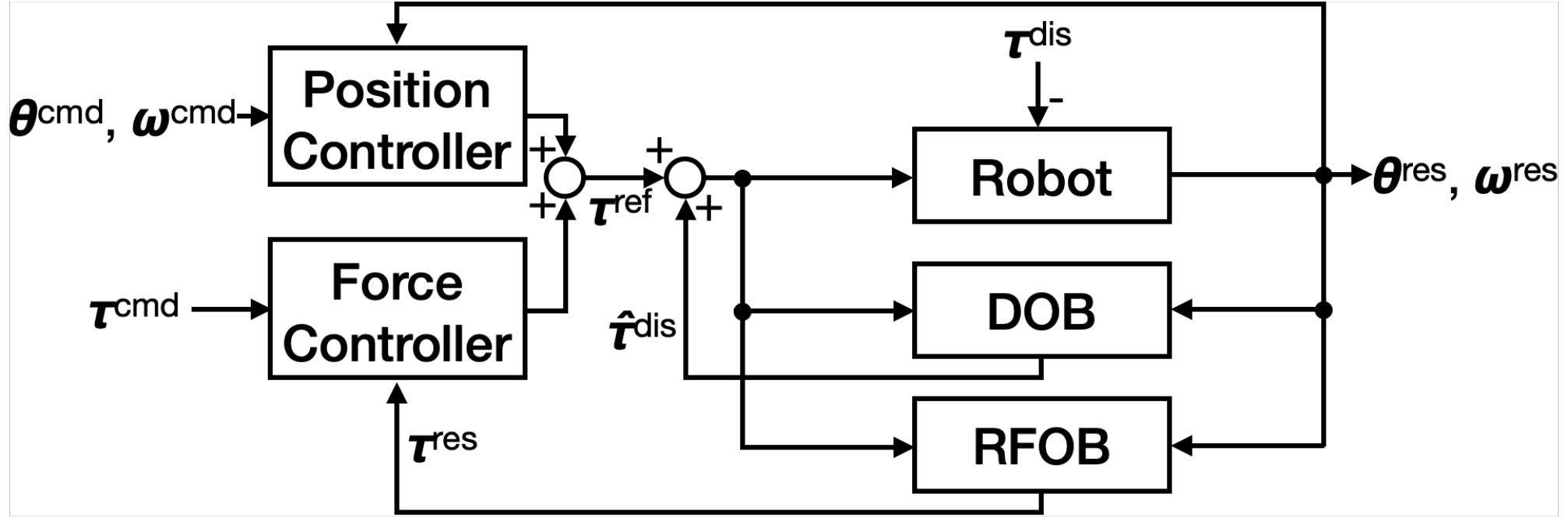}}
                \caption{Block diagram of the controller for each joint of the manipulators.}
                \label{fig:controller}
            \end{figure}

            To enable torque sensing without torque sensors, each joint of the manipulator was controlled with a disturbance observer~(DOB) and reaction force observer~(RFOB), as shown in Fig.~\ref{fig:controller}, with parameters identical to those used by Saigusa \textit{et al.}~\cite{CRANEX7params}.
            The variables ${\theta}$, ${\omega}$, and ${\tau}$ denote the joint angle, angular velocity, and torque, respectively.
            The superscripts $cmd$, $ref$, $res$, and $dis$ denote command, reference, response, and disturbance, respectively.
            The circumflexes~(~$\hat{}$~) denote the DOB estimates.
            
        \subsubsection{Setup of Learning Models}
\begin{figure}[tb]
                \centerline{\includegraphics[scale=0.28, bb=0 0 930 308]{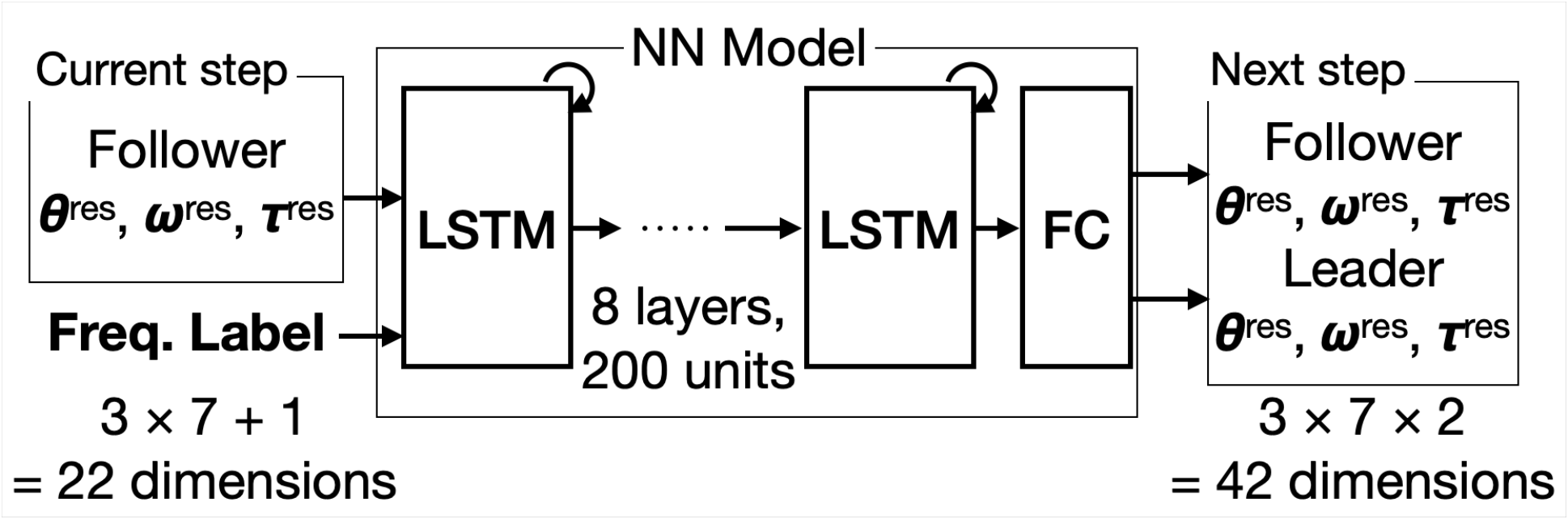}}
                \caption{Configuration of the NN model.}
                \label{fig:nnmodel}
            \end{figure}

            The neural network (NN) model shown in Fig.~\ref{fig:nnmodel} consisted of eight long short-term memory (LSTM)~\cite{LSTM1, LSTM2} layers with 200 dimensions and a fully connected (FC) layer before the output layer.
            The model had 22 dimensions, including the follower's joint angles, angular velocities, and torques, in addition to the real-world frequency of the wiping motion.
            The outputs had 42 dimensions consisting of both the follower's and leader's states.
            The configuration of the inputs and outputs was based on the follower-to-follower/leader setting~\cite{F2FL, akagawaJIA}, enabling the prediction of both the command and environmental reaction and making validation through autoregression possible.
            The variables ${\theta}$, ${\omega}$, and ${\tau}$ denote the joint angle, angular velocity, and torque, respectively.
            The superscripts $cmd$, $ref$, $res$ and $dis$ denote command, reference, response, and disturbance, respectively.

            The sampling frequency of the NN model, $F$, was set to 25~Hz.
            For the training data without VFIL, all the 500~Hz demonstration data were downsampled to 20 episodes each~\cite{yamaneral, rahmatizadeh2018}.
            For the training data with VFIL, $f_0$ was set to 0.6~Hz, and the data with a wiping frequency $f$ of 0.6~Hz were treated in the same manner as those without VFIL. 
            Wiping data with $f$ values of 0.4~Hz and 0.8~Hz were first resampled to 333.3~Hz and 666.7~Hz using linear interpolation and then downsampled to 20 episodes each, with the sampling frequency $fF/f_0$ at 16.7~Hz and 33.3~Hz, respectively.

    \subsection{Results}
        Table~\ref{vfilcfil} shows the success rates for the wiping task. 
        The proposed method achieved higher success rates than the conventional method for both high- and low-height settings, contributing to a 12.5~\% increase in the total success rate.
        All failures were due to the loss of contact during the wiping motion in the low-height setting.
        This is because loss of contact is unlikely when the surface is closer to the robot than predicted, and the NN models have learned to adapt to such closeness.
        In all failed trials, the wiping motion continued even after the loss of contact, and the contacts were recovered.

        \begin{table}[tb]
            \caption{Success Rate}
            \begin{center}
                \subfloat[With Variable-Frequency Imitation Learning]{
                    \resizebox*{\hsize}{!}{
                        \begin{tabular}{|c|c|c|c|c|c|}
                            \hline
                            \textbf{Height}&\multicolumn{4}{|c|}{\textbf{Frequency Label (Hz)}}& \\
                            \cline{2-5} 
                            \textbf{(cm)} & \textbf{\textit{0.2}}& \textbf{\textit{0.6}}& \textbf{\textit{1.0}} & \textbf{\textit{1.4}}  & \textbf{\textit{Total}} \\
                            \hline
                            15 & \emph{5/5(100\%)} &  \emph{5/5(100\%)} &  \emph{5/5(100\%)} &  \emph{5/5(100\%)} &  \emph{20/20(100\%)}  \\
                            \hline
                            10 & 1/5(20\%) &  \emph{3/5(60\%)} &  \emph{5/5(80\%)} &  \emph{5/5(100\%)} &  \emph{14/20(70\%)}  \\
                            \hline
                            Total & 6/10(60\%) &  \emph{8/10(80\%)} &  \emph{10/10(100\%)} &  \emph{10/10(100\%)} &  \emph{34/40(85\%)}  \\
                            \hline
                        \end{tabular}
                    }
                \label{vfil}
                }
            \end{center}
            \vspace{5pt}
            \begin{center}
                \subfloat[Without Variable-Frequency Imitation Learning]{
                    \resizebox*{\hsize}{!}{
                        \begin{tabular}{|c|c|c|c|c|c|}
                            \hline
                            \textbf{Height}&\multicolumn{4}{|c|}{\textbf{Frequency Label (Hz)}}& \\
                            \cline{2-5} 
                            \textbf{(cm)} & \textbf{\textit{0.2}}& \textbf{\textit{0.6}}& \textbf{\textit{1.0}} & \textbf{\textit{1.4}}  & \textbf{\textit{Total}} \\
                            \hline
                            15 & \emph{5/5(100\%)} &  \emph{5/5(100\%)} &  \emph{5/5(100\%)} &  \emph{5/5(100\%)} &  \emph{20/20(100\%)}  \\
                            \hline
                            10 & \emph{3/5(60\%)} &  \emph{3/5(60\%)} &  1/5(20\%) &  2/5(40\%) &  9/20(45\%)  \\
                            \hline
                            Total & \emph{8/10(80\%)} &  \emph{8/10(80\%)} &  6/10(60\%) &  7/10(70\%) &  29/40(73\%)  \\
                            \hline
                        \end{tabular}
                    }
                \label{cfil}
                }
            \end{center}
            \label{vfilcfil}
        \end{table}

\begin{figure}[tb]
            \centerline{\includegraphics[scale=0.7, bb=0 0 360 219]{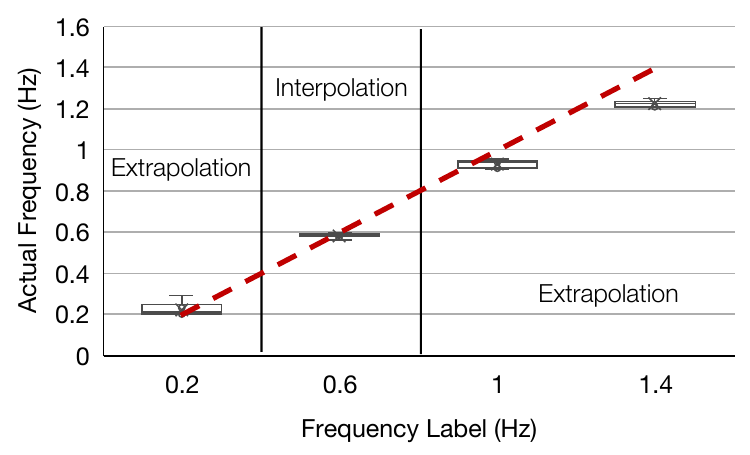}}
            \caption{Actual frequencies of successful trials with VFIL. Data being closer to the dashed line indicates a better fit.}
            \label{fig:vfil12}
        \end{figure}

\begin{figure}[tb]
            \centerline{\includegraphics[scale=0.7, bb=0 0 360 219]{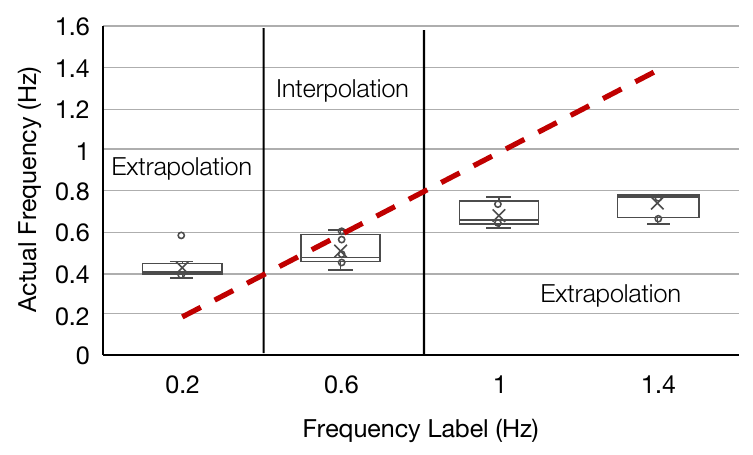}}
            \caption{Actual frequency of successful trials without VFIL. Data being closer to the dashed line indicates a better fit.}
            \label{fig:cfil40}
        \end{figure}

        Figs.~\ref{fig:vfil12}~and~\ref{fig:cfil40} depict the actual frequency of the wiping motion in successful trials with and without VFIL. 
        The proposed method contributed to a smaller variance from the frequency label, even at untaught frequencies, whereas the conventional method failed to perform outside the frequencies in the teaching data, that is, between 0.4 and 0.8~Hz.
        The larger variance at 1.4~Hz with VFIL was likely due to increased friction and other nonlinearities in the robot system, indicating that the success of the task was prioritized over the frequencies of the motions.

    \subsection{Discussions}
        Improvements in both the success rate and accuracy suggest that the proposed method succeeded in generating variable-speed motion, even outside the range of training data.
        In the high-frequency motion, the success of the task was prioritized over moving quickly and failing, although there were some delays.
        This prioritization is unlikely in simple playbacks, and delays can be overcome by fine-tuning using self-supervised learning~\cite{CRANEX7params}.
        The increased number of failures in the low-frequency setting is likely due to the increased sampling time of the NN model, which reached 120~ms at a wiping frequency $f =$~0.2~Hz.

        Below describes the limitations of the proposed method.

        First, in this method, there is a tradeoff between the feasibilities of faster and slower motions.
        This is because the interval of the NN model is given solely by $f_0/fF$, and because of the complexity of the NN model, certain calculation time is required at each step.
        A smaller sampling frequency $F$ enables higher-frequency motion with a larger $f/f_0$, which leads to a longer $f_0/fF$ and a jaggier motion at a lower motion frequency $f$.
        Training data with different base frequencies $f_0$ may enable both, although this leads to multi-task learning settings and may affect the success rate and accuracy.
        The interpolation of samples in a predictive manner can resolve this issue.

        Second, because this method requires a frequency label prior to the inference phase, the frequency of motion does not change autonomously.
        Our future work will include the use of a hierarchical architecture~\cite{kaisogata} to determine the velocity of movement according to the situation.

\section{Conclusion}
    This paper proposes variable-frequency imitation learning, a novel imitation learning method with time-wise normalization that enables inference at variable frequencies according to the frequency of motion.
    By adding this inductive bias, the learning model tasks are reduced by generating a variable-speed motion to adapt to the additional nonlinearity. 

    The experimental results of the wiping task show a remarkable improvement in accuracy along the frequency label and an increased success rate in high-frequency settings. 
    Our future work will involve widening the frequency range and enabling autonomous decisions regarding the frequency of motion.

\bibliographystyle{IEEEtran}
\bibliography{masuya.bib}

% Generated by IEEEtran.bst, version: 1.14 (2015/08/26)
\begin{thebibliography}{10}
\providecommand{\url}[1]{#1}
\csname url@samestyle\endcsname
\providecommand{\newblock}{\relax}
\providecommand{\bibinfo}[2]{#2}
\providecommand{\BIBentrySTDinterwordspacing}{\spaceskip=0pt\relax}
\providecommand{\BIBentryALTinterwordstretchfactor}{4}
\providecommand{\BIBentryALTinterwordspacing}{\spaceskip=\fontdimen2\font plus
\BIBentryALTinterwordstretchfactor\fontdimen3\font minus
  \fontdimen4\font\relax}
\providecommand{\BIBforeignlanguage}[2]{{%
\expandafter\ifx\csname l@#1\endcsname\relax
\typeout{** WARNING: IEEEtran.bst: No hyphenation pattern has been}%
\typeout{** loaded for the language `#1'. Using the pattern for}%
\typeout{** the default language instead.}%
\else
\language=\csname l@#1\endcsname
\fi
#2}}
\providecommand{\BIBdecl}{\relax}
\BIBdecl

\bibitem{levine2018}
S.~Levine, P.~Pastor, A.~Krizhevsky, J.~Ibarz, and D.~Quillen, ``{Learning
  Hand-Eye Coordination for Robotic Grasping with Deep Learning and Large-Scale
  Data Collection},'' \emph{The International Journal of Robotics Research},
  vol.~37, no. 4-5, pp. 421--436, 2018.

\bibitem{daydreamer}
P.~Wu, A.~Escontrela, D.~Hafner, P.~Abbeel, and K.~Goldberg, ``Daydreamer:
  World models for physical robot learning,'' in \emph{Proceedings of The 6th
  Conference on Robot Learning}, ser. Proceedings of Machine Learning Research,
  K.~Liu, D.~Kulic, and J.~Ichnowski, Eds., vol. 205.\hskip 1em plus 0.5em
  minus 0.4em\relax PMLR, 14--18 Dec 2023, pp. 2226--2240.

\bibitem{yang2017}
P.-C. Yang, K.~Sasaki, K.~Suzuki, K.~Kase, S.~Sugano, and T.~Ogata,
  ``{Repeatable Folding Task by Humanoid Robot Worker Using Deep Learning},''
  \emph{IEEE Robotics and Automation Letters}, vol.~2, no.~2, pp. 397--403,
  2017.

\bibitem{aloha}
T.~Z. Zhao, V.~Kumar, S.~Levine, and C.~Finn, ``{Learning Fine-Grained Bimanual
  Manipulation with Low-Cost Hardware},'' in \emph{Proceedings of Robotics:
  Science and Systems}, Daegu, Republic of Korea, July 2023.

\bibitem{guan2024}
Y.~Guan, H.~Liao, Z.~Li, J.~Hu, R.~Yuan, Y.~Li, G.~Zhang, and C.~Xu, ``World
  models for autonomous driving: An initial survey,'' \emph{IEEE Transactions
  on Intelligent Vehicles}, pp. 1--17, 2024.

\bibitem{mitrokhov2024}
K.~Mitrokhov, ``Between world models and model worlds: on generality, agency,
  and worlding in machine learning,'' \emph{AI \& SOCIETY}, 2024.

\bibitem{cohen2021}
S.~Cohen, G.~Luise, A.~Terenin, B.~Amos, and M.~Deisenroth, ``Aligning time
  series on incomparable spaces,'' in \emph{Proceedings of The 24th
  International Conference on Artificial Intelligence and Statistics}, ser.
  Proceedings of Machine Learning Research, A.~Banerjee and K.~Fukumizu, Eds.,
  vol. 130.\hskip 1em plus 0.5em minus 0.4em\relax PMLR, 13--15 Apr 2021, pp.
  1036--1044.

\bibitem{sakoe1978}
H.~Sakoe, ``Dynamic programming algorithm optimization for spoken word
  recognition,'' \emph{IEEE Transactions on Acoustics, Speech, and Signal
  Processing}, vol.~26, pp. 159--165, 1978.

\bibitem{rasines2023}
I.~Rasines, A.~Remazeilles, M.~Prada, and I.~Cabanes, ``Minimum cost averaging
  for multivariate time series using constrained dynamic time warping: A case
  study in robotics,'' \emph{IEEE Access}, vol.~11, pp. 80\,600--80\,612, 2023.

\bibitem{saveriano2023}
M.~Saveriano, F.~J. Abu-Dakka, A.~Kramberger, and L.~Peternel, ``Dynamic
  movement primitives in robotics: A tutorial survey,'' \emph{The International
  Journal of Robotics Research}, vol.~42, no.~13, pp. 1133--1184, 2023.

\bibitem{XU2024}
X.~Xu, K.~Qian, B.~Zhou, F.~Fang, and X.~Ma, ``Imitating via manipulability:
  Geometry-aware combined dmp with via-point and speed adaptation,''
  \emph{Computers and Electrical Engineering}, vol. 117, p. 109247, 2024.

\bibitem{perico2020}
C.~V. Perico, J.~de~Schutter, and E.~Aertbeli^^c3^^abn, ``Learning robust
  manipulation tasks involving contact using trajectory parameterized
  probabilistic principal component analysis,'' in \emph{2020 IEEE/RSJ
  International Conference on Intelligent Robots and Systems (IROS)}, 2020, pp.
  8336--8343.

\bibitem{perico2019}
C.~A.~V. Perico, J.~De~Schutter, and E.~Aertbeli^^c3^^abn, ``Combining
  imitation learning with constraint-based task specification and control,''
  \emph{IEEE Robotics and Automation Letters}, vol.~4, no.~2, pp. 1892--1899,
  2019.

\bibitem{yokokura2009}
Y.~Yokokura, S.~Katsura, and K.~Ohishi, ``Stability analysis and experimental
  validation of a motion-copying system,'' \emph{IEEE Transactions on
  Industrial Electronics}, vol.~56, no.~10, pp. 3906--3913, 2009.

\bibitem{fujisaki2023}
K.~Fujisaki and S.~Katsura, ``Motion-copying system with in-tool sensing,''
  \emph{IEEJ Journal of Industry Applications}, vol.~12, no.~4, pp. 793--799,
  2023.

\bibitem{sakaino2022}
S.~Sakaino, K.~Fujimoto, Y.~Saigusa, and T.~Tsuji, ``Imitation learning for
  variable speed contact motion for operation up to control bandwidth,''
  \emph{IEEE Open Journal of the Industrial Electronics Society}, vol.~3, pp.
  116--127, 2022.

\bibitem{CRANEX7params}
Y.~Saigusa, S.~Sakaino, and T.~Tsuji, ``{Imitation Learning for Nonprehensile
  Manipulation Through Self-Supervised Learning Considering Motion Speed},''
  \emph{IEEE Access}, vol.~10, pp. 68\,291--68\,306, 2022.

\bibitem{yamamoto2023}
K.~Yamamoto, H.~Ito, H.~Ichiwara, H.~Mori, and T.~Ogata, ``Real-time motion
  generation and data augmentation for grasping moving objects with dynamic
  speed and position changes,'' in \emph{2024 IEEE/SICE International Symposium
  on System Integration (SII)}, 2024, pp. 390--397.

\bibitem{adachi2018}
T.~Adachi, K.~Fujimoto, S.~Sakaino, and T.~Tsuji, ``{Imitation Learning for
  Object Manipulation Based on Position/Force Information Using Bilateral
  Control},'' in \emph{2018 IEEE/RSJ International Conference on Intelligent
  Robots and Systems (IROS)}, 2018, pp. 3648--3653.

\bibitem{sakaino2011}
S.~Sakaino, T.~Sato, and K.~Ohnishi, ``{Multi-DOF Micro-Macro Bilateral
  Controller Using Oblique Coordinate Control},'' \emph{IEEE Transactions on
  Industrial Informatics}, vol.~7, no.~3, pp. 446--454, 2011.

\bibitem{yamaneral}
K.~Yamane, Y.~Saigusa, S.~Sakaino, and T.~Tsuji, ``Soft and rigid object
  grasping with cross-structure hand using bilateral control-based imitation
  learning,'' \emph{IEEE Robotics and Automation Letters}, vol.~9, no.~2, p.
  1198^^e2^^80^^931205, Feb. 2024.

\bibitem{LSTM1}
S.~Hochreiter and J.~Schmidhuber, ``{Long Short-Term Memory},'' \emph{Neural
  computation}, vol.~9, pp. 1735--80, 12 1997.

\bibitem{LSTM2}
H.~Sak, A.~W. Senior, and F.~Beaufays, ``{Long Short-Term Memory Recurrent
  Neural Network Architectures for Large Scale Acoustic Modeling},'' in
  \emph{Interspeech}, 2014, pp. 338--342.

\bibitem{F2FL}
K.~Hayashi, A.~Sasagawa, S.~Sakaino, and T.~Tsuji, ``{A New Autoregressive
  Neural Network Model with Command Compensation for Imitation Learning Based
  on Bilateral Control},'' in \emph{2021 IEEE International Conference on
  Mechatronics (ICM)}, 2021, pp. 1--7.

\bibitem{akagawaJIA}
T.~Akagawa and S.~Sakaino, ``Autoregressive model considering low frequency
  errors in command for bilateral control-based imitation learning,''
  \emph{IEEJ Journal of Industry Applications}, vol.~12, no.~1, pp. 26--32,
  2023.

\bibitem{rahmatizadeh2018}
R.~Rahmatizadeh, P.~Abolghasemi, A.~Behal, and L.~B^^c3^^b6l^^c3^^b6ni, ``From
  virtual demonstration to real-world manipulation using lstm and mdn,''
  \emph{Proceedings of the AAAI Conference on Artificial Intelligence},
  vol.~32, no.~1, Apr. 2018.

\bibitem{kaisogata}
K.~Hayashi, S.~Sakaino, and T.~Tsuji, ``{An Independently Learnable
  Hierarchical Model for Bilateral Control-Based Imitation Learning
  Applications},'' \emph{IEEE Access}, vol.~10, pp. 32\,766--32\,781, 2022.

\end{thebibliography}

\end{document}